\documentclass[sigconf,nonacm]{acmart}
\usepackage{listings}

\AtBeginDocument{%
  \providecommand\BibTeX{{%
    \normalfont B\kern-0.5em{\scshape i\kern-0.25em b}\kern-0.8em\TeX}}}

\setcopyright{none}
\renewcommand\footnotetextcopyrightpermission[1]{}
\setcopyright{none}
\copyrightyear{2023}
\acmYear{}
\acmDOI{}

\acmPrice{}
\acmISBN{}

\begin{document}

\title[Supermind Ideator]{Supermind Ideator:\\ Exploring generative AI to support creative problem-solving}

\author{Steven R. Rick $^1$}
\authornote{Both authors contributed equally to this research.}
\author{Gianni Giacomelli $^1$}
\authornotemark[1]
\author{Haoran Wen $^1$}
\author{Robert J. Laubacher $^1$}
\author{Nancy Taubenslag $^1$}
\author{Jennifer L. Heyman $^1$}
\affiliation{%
  \institution{$^1$Center for Collective Intelligence \\ Massachusetts Institute of Technology}
  \country{USA}
}
\author{Max Sina Knicker $^1$}
\author{Younes Jeddi $^1$ $^2$ }
\author{Hendrik Maier $^1$}
\author{Stephen Dwyer $^1$}
\author{Pranav Ragupathy $^1$}
\author{Thomas W. Malone $^1$}
\affiliation{%
  \institution{$^2$School of Collective Intelligence \\ Mohammed VI Polytechnic University}
  \country{Morocco}
}

\renewcommand{\shortauthors}{Rick, Giacomelli, et al.}

\begin{abstract}
  Previous efforts to support creative problem-solving have included (a) techniques (such as brainstorming and design thinking) to stimulate creative ideas, and (b) software tools to record and share these ideas. Now, generative AI technologies can suggest new ideas that might never have occurred to the users, and users can then select from these ideas or use them to stimulate even more ideas. Here, we describe such a system, Supermind Ideator. The system uses a large language model (GPT 3.5) and adds prompting, fine tuning, and a user interface specifically designed to help people use creative problem-solving techniques. Some of these techniques can be applied to any problem; others are specifically intended to help generate innovative ideas about how to design groups of people and/or computers (“superminds”). We also describe our early experiences with using this system and suggest ways it could be extended to support additional techniques for other specific problem-solving domains.
\end{abstract}


\maketitle

\section{Introduction}

Creative problem-solving is critical to success in many kinds of human activity, from architecture, engineering, and software development to art, entrepreneurship, and designing the human groups that perform all these activities \cite{mednick_associative_1963, dorst_creativity_2001, gabora_cognitive_2002, sanders_co-creation_2008, cardoso_fixation_2011}. It is, therefore, not surprising that many techniques to improve creative problem-solving have been proposed over the years, including brainstorming, design thinking, mind-mapping, crowdsourcing, and many others \cite{robertson_role_2008, frich_twenty_2018, griebel_augmented_2020}.
 
In this paper, we investigate the potential of a new kind of tool--generative AI--for supporting creative problem-solving. In particular, we focus on how large language models (LLMs; e.g., GPT (Generative Pre-trained Transformer) \cite{brown_language_2020}) can take natural language descriptions of a problem as input and produce as output natural language ideas about how to reframe or solve the problem.

Interestingly, even though one widely discussed limitation of today's LLMs is that they sometimes produce incorrect or irrelevant outputs, this limitation is usually not a problem when the generative AI system is used to \textit{augment} human creativity instead of \textit{replacing} it. In this case, human users can often easily decide which of the outputs from the system are useful enough to consider further and which aren't. And even ideas that may at first seem irrelevant can sometimes trigger further useful ideas for human users. In fact, trying to make connections between a problem and seemingly unrelated ideas is one simple technique for triggering creative ideas \cite{lee_prompts_2023}.

To do this, we developed an LLM-based system, called Supermind Ideator, that uses specialized prompts, fine-tuning, and a user interface to generate ideas that help users reflect upon their problems and generate possible solutions. The system does this using a set of conceptual \textit{moves}--techniques that humans can use to trigger creative ideas. By sequencing these moves in different orders and combinations, users can explore many different ideas for a given problem.

Most of the techniques we currently use in the Supermind Ideator are based on the "Supermind Design" methodology \cite{noauthor_supermind_nodate}. Some of these techniques, such as looking at sub-parts or analogies, can be helpful for addressing any problem. Other techniques are specifically intended to help generate innovative ideas about how to design \textit{superminds}, defined as \textit{groups of individuals acting together in ways that seem intelligent} \cite{malone2018superminds}. For instance, one such supermind design technique encourages users to consider how a problem could be solved with different kinds of groups, such as hierarchies, democracies, markets, or communities.

In other words, "superminds" is a short way of saying "collectively intelligent systems," and the rest of this paper is about how the Supermind Ideator uses the Supermind Design methodology to help design such systems. 

\section{Related Work}

Prior work has extensively discussed creative problem-solving approaches that use techniques from design thinking to collective intelligence. These approaches predominantly focus on groups using organizational and methodological approaches to address issues such as design fixation, knowledge curation, and creative inspiration. 

\subsection{Facilitated Idea Generation}

\begin{figure}[h]
    \centering
    \includegraphics[width=\linewidth]{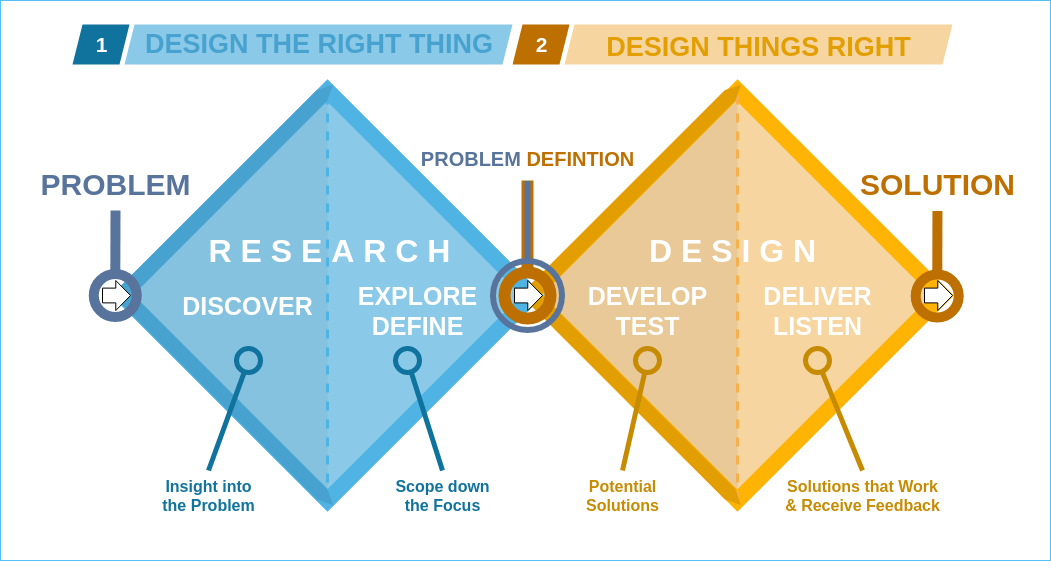}
    \caption{The Double Diamond method which highlights the two diamonds (problem and solution spaces) and showcases the four steps where you first discover and define the problem before then developing and delivering the solution. Source: Wikipedia - Double Diamond (design process model)}
    \Description{The Double Diamond method which highlights the two diamonds (problem and solution spaces) and showcases the four steps where you first discover and define the problem before then developing and delivering the solution. Source: Wikipedia - Double Diamond (design process model)}
    \label{fig:double_diamond}
\end{figure}

From the early days of design being applied in the scientific arena \cite{cross_designerly_1982} to the modern-day frameworks for using Design Thinking methodologies \cite{dam_design_2020}, creative problem-solving techniques have evolved significantly. For example, the mix of deep problem understanding and iterative solution generation, most notably combined in the Double Diamond \cite{noauthor_framework_2005}, enables a rigorous and empathetic approach to integrate the needs of people, the possibilities of technology, and the requirements for business success. In the first of the two diamonds, practitioners (a) \textit{diverge} by considering different ways to frame the problem and then (b) \textit{converge} to narrow down to a useful problem definition. Then, in the second diamond, they (a) \textit{diverge} by considering different potential solutions to the problem and then (b) \textit{converge} on a few of the best solutions, as shown in Fig \ref{fig:double_diamond}.

Other prior work has studied the phenomenon of creative ideation through sociocultural lenses \cite{dorst_creativity_2001, lubart_models_2001, gabora_cognitive_2002, pavie_leveraging_2015, frich_twenty_2018}, digging into specific topics and issues such as design fixation \cite{linsey_study_2010, cardoso_fixation_2011, smith_three-pronged_2011, youmans_design_2014}, bias \cite{mednick_associative_1963}, inspiration \cite{eckert_sources_2000, thrash_psychology_2014}, and innovation \cite{howard_use_2010, pavie_leveraging_2015}. According to Gabora, for instance, ideas emerge, evolve, and manifest as creative products through the formation of combinations and reorganization of existing ideas \cite{gabora_cognitive_2002}. The process can be modeled by constraint-based iteration to transform ideas into tangible solutions. 

However, idea-generation activities can be long and laborious, and mental tendencies, such as functional fixedness, often lead to design fixation \cite{jansson_design_1991}. Design fixation is a cognitive bias that presents as an over-fitting of design space to pre-existing knowledge and experiences of the designer \cite{cardoso_fixation_2011}. It hinders wide-reaching ideas and overly constrains the domain of idea generation, functionally falling into a local maximum and missing the wider absolute maximum available. Studies have attempted to develop prevention methods such as concept mapping, remote association, facilitated design thinking, and exposure to other new and unrelated content \cite{howard_use_2010, smith_three-pronged_2011, youmans_design_2014}. More recent work has examined the potential for collective intelligence approaches such as crowdsourcing idea generation, crowd-based ratings, and other forms of computer-supported cooperative work to help alleviate design fixation and access diverse types of knowledge \cite{gregg_designing_2010, majchrzak_towards_2013, lee_how_2019, yun_collective_2021, klein2014embarrassment}. All of these approaches, however, still largely rely on human effort and access to knowledge resources that do not easily scale.

\subsection{Generative AI}

The rise of Transformer architectures in language modeling has given a meteoric rise to the accessibility of LLMs \cite{vaswani_attention_2017}.  New systems like GPT (Generative Pre-trained Transformer) have furthered this to give more general access to inferencing models that rapidly produce large-scale human-quality text output from a small input \cite{brown_language_2020}. These models can be tuned and directed with either (a) simple instructions (\textit{zero-shot learning}), (b) small numbers of example input and output pairs (\textit{few-shot learning}), or (c) large numbers of example inputs and outputs (\textit{fine-tuning}). With these directions, it has been shown that LLMs are capable of very wide-ranging idea generation \cite{summers-stay_brainstorm_2023}, and increasing work is being done to understand how to improve the quality of output from LLMs to better match the design space constraints \cite{zhu_generative_2023}. Prior research has told us that generating more ideas can lead to greater creativity \cite{paulus2011effects} so it seems plausible that using generative AI might be able to help us to generate more ideas and in turn lead to greater creativity.

To our knowledge, little systematic work has explicitly explored how to combine the expertise of human designers with the capabilities of generative AI technologies. But given that current generative AI systems by themselves are often not capable of producing uniformly high-quality ideas, a very promising possibility is to let generative AI systems rapidly generate ideas for humans to consider. Generative technologies are already more than capable enough to provide usefully unexpected inputs and stimuli for human designers. And this can increase the range of possibilities to be considered by humans. Thoughtfully designing the way a system guides and facilitates this process can thus improve the probability that designers will find better solutions faster than they would have otherwise. 

Two core facets underlie this concept: \textit{reflection} and \textit{inspiration}.

\subsubsection{Reflection}

During creative problem-solving, it is often useful to engage in a reflective feedback loop \cite{wooten_idea_2017, chen_design_2023}. While random feedback can be much better than no feedback, an approach that facilitates reflection can prevent early fixation on solutions and help explore design space constraints from both the problem and solution domains \cite{nakata_design_2020}. 

We extend this perspective by observing that much of the work done by design thinking and innovation facilitators is less an act of solution generation and curation and more of prompting insight through introspective questions and reflective actions. With this in mind, we suggest a useful characteristic for any innovation support system - it should promote reflection by humans rather than impose unnecessary constraints on their thinking. For instance, considering system outputs as "ideas" rather than "answers" and positioning the system as a tool to \textit{support} human cognition rather than \textit{replace} it may be key. And it seems especially important to emphasize to the human users that they are responsible for deciding what to do with the system’s outputs. They can ignore the system outputs, use them without change, or let them stimulate further creative thinking by humans.

\subsubsection{Inspiration}

The second critical quality for any innovation support system is to promote wide-reaching exploration of the available design space. This is often accomplished through facilitated activities that promote divergent rather than convergent thinking, such as the alternative uses test \cite{gilhooly_divergent_2007}. Many divergent thinking activities put the onus of creating new uses on the human engaging in the activity \cite{bygstad_generative_2010}. Fortunately, it is increasingly being found that generative technologies like GPT can do quite well to replicate human idea generation within such divergent thinking tasks \cite{summers-stay_brainstorm_2023}. This does not replace human ideation. Instead, it presents a chance to augment human idea generation and avoid design fixation by exposing humans to novel and diverse stimuli that stimulate new connections \cite{howard_use_2010, smith_three-pronged_2011, youmans_design_2014}. 

A complex topic concerning LLMs today is their safety and accuracy with generated content \cite{bender_dangers_2021, singh_where_2022}. Transformer-based systems are capable of exceptionally human-like language generation, but they do not always produce factual content. These "hallucinations" by the machine can sound as if they carry a sense of authority and accuracy even when they are simply false combinations of words that statistically sound appropriate given prior text and the underlying model of the LLM \cite{goodman_lampost_2022}. 

We see this (and much of the rest of the content that LLMs can create) more as an opportunity and feature rather than a bug. Outlandish and unfeasible ideas are a natural fit for idea-generation practices. By interrogating the extremes of a design space, it is possible to construct a better understanding of what is actually possible and appropriate. By framing all content as ideas (including both good and bad ideas), we present the designer with a more far-reaching design space exploration. Such support for divergent thinking is essential for the success of any creative problem-solving support system.

\section{The Supermind Design Methodology}

The Supermind Ideator is based on the Supermind Design methodology, which includes a set of conceptual moves that people can use to spur their creativity about how to design collectively intelligent groups \cite{noauthor_supermind_nodate}. These moves have been used successfully in multiple settings \cite{koppineni2022supermind, laubacher_using_2020}. 

The methodology includes the following \textit{basic design moves} that are based on general techniques for any kind of creative problem-solving \cite{osborn2012applied, de1970lateral, koberg1974universal, brown_2009, dschool_design_bootleg, malone2003organizing}: 
\begin{itemize}
    \item \textit{Zoom In - Parts}: What are the parts of this problem?
    \item \textit{Zoom In - Types}: What are the types of this problem?
    \item \textit{Zoom Out - Parts}: What is this problem a part of?
    \item \textit{Zoom Out - Types}: What is this problem a type of?
    \item \textit{Analogize}: What are analogies for this problem?   
\end{itemize}

The methodology also includes the following \textit{supermind design moves} that are specifically for generating ideas about how to design superminds (i.e., collectively intelligent groups) \cite{malone2018superminds, noauthor_supermind_nodate}: 
\begin{itemize}
    \item \textit{Groupify}: How can different kinds of groups help solve the problem? Possibilities include:
    \begin{itemize}
        \item \textit{Hierarchy} - where group decisions are made by delegating them to individuals who possess varying levels of authority within an organization
        \item \textit{Democracy} - where group decisions are made by voting
        \item \textit{Market} - where group decisions are the combination of all the pairwise agreements between individual buyers and sellers
        \item \textit{Community} - where group decisions are made by informal consensus based on shared norms and reputations
        \item \textit{Ecosystem} - where group decisions are made by whoever has the most power and by survival of the fittest
    \end{itemize}
    \item \textit{Cognify}: How can different cognitive processes help solve the problem? Possibilities include:
    \begin{itemize}
        \item \textit{Create} - How can groups create new information collectively?
        \item \textit{Decide} - How can groups make choices?
        \item \textit{Sense} - How can groups collect and interpret information from the environment?
        \item \textit{Remember} - How can groups recall information from the past?
        \item \textit{Learn} - How can groups improve their performance with experience?
    \end{itemize}
    \item \textit{Technify}: How can technologies be used to help solve the problem?
\end{itemize}

\begin{figure*}[h!]
  \centering
  \includegraphics[width=\textwidth]{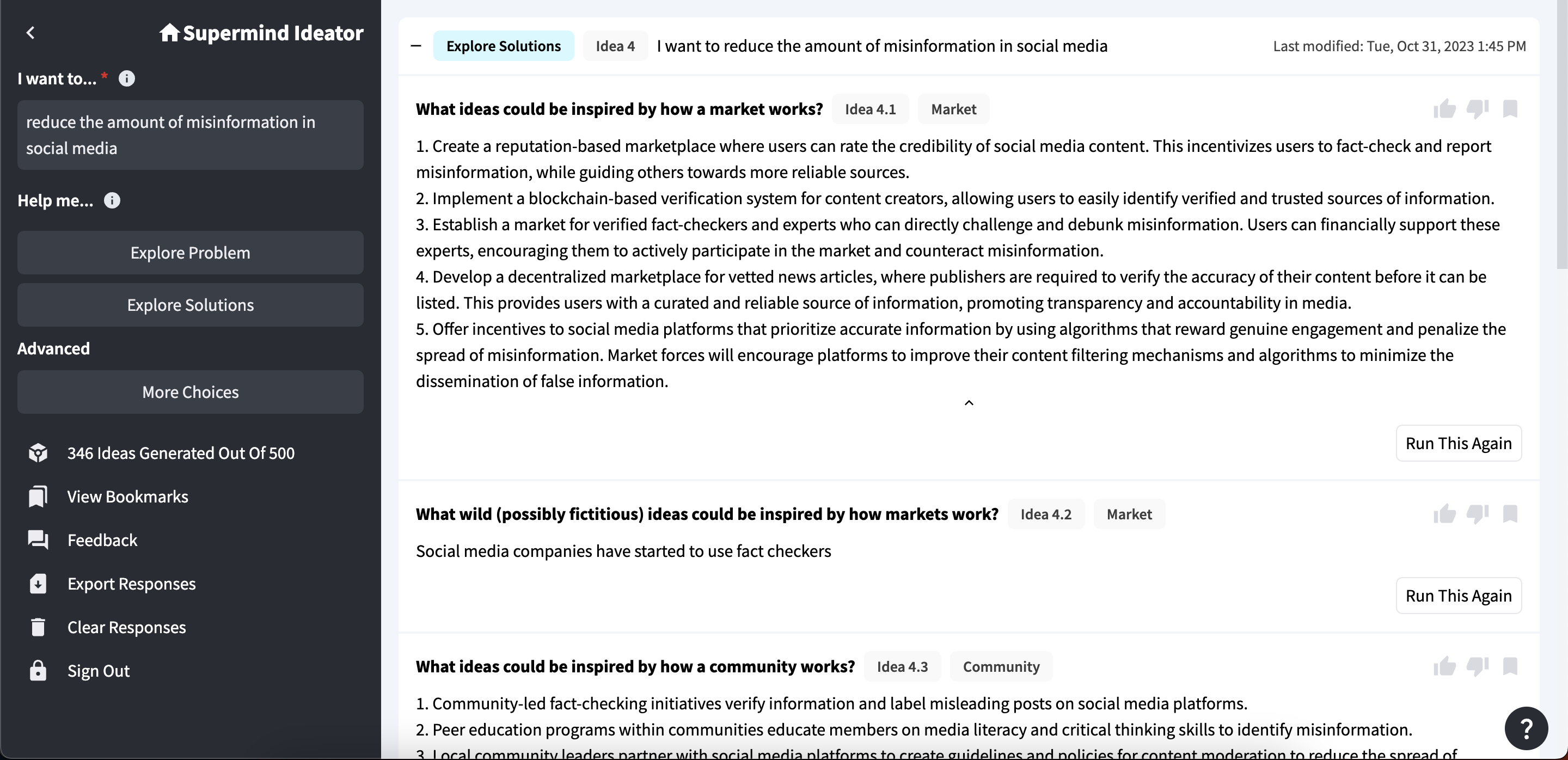}
  \caption{The Supermind Ideator Interface. The left  side contains the Generate Panel where users input their problem and select Moves to run. The right side contains ideas generated by the system.}
  \Description{The Supermind Ideator Interface. The left  side contains the Generate Panel where users input their problem and select Moves to run. The right side contains ideas generated by the system.}
  \label{fig:overallIdeator}
\end{figure*}

Finally, the methodology includes three \textit{experimental moves} which have not, to our knowledge, been previously used as part of systematic ideation exercises but which appear to take advantage of GPT's capabilities and are incorporated in the Supermind Ideator:
\begin{itemize}
    \item \textit{Reflect} - What is missing from the current problem statement?
    \item \textit{Reformulate} - How could the problem be reformulated?
    \item \textit{Case examples} - How does the problem relate to case examples of real companies and products?
\end{itemize}

\section{The Supermind Ideator Software}

The Supermind Ideator \footnote{https://ideator.mit.edu/} was designed to guide users into a focused state of idea generation and reflection, and as such is intentionally kept minimalistic, shown in Fig \ref{fig:overallIdeator}. After users type in their problem, they are given three options: \textit{Explore Problem}, \textit{Explore Solutions}, and \textit{More Choices}. These options are meant to provide scaffolding for novice users who may not know where to begin, as well as to support more advanced users who already know what they want to do next. The first two options, Explore Problems and Explore Solutions, comprise what we have called "move sets," or groups of moves that focus on a specific aspect of the idea generation and refinement process. 

The Explore Problem move set supports the problem definition part of the double diamond approach using the \textit{basic design moves} and the \textit{experimental moves} from the Supermind Design methodology. In this way, it helps users reflect on how they can generalize and specialize the various parts and types of their problem, consider relevant analogies to their problem, and identify potentially missing aspects of their problem statement. 

The Explore Solutions move set supports the solution generation part of the double diamond approach using the \textit{supermind design moves} from the Supermind Design methodology. For example, it helps users consider how different kinds of groups (such as markets, communities, and democracies) could help solve their problem. It also helps users think about how innovative ways of performing different cognitive processes (such as creating, deciding, and sensing) or using various kinds of technologies could help solve their problem.

More Choices allows a user to select any individual move(s) they want. The More Choices option also exposes a more advanced parameter that the GPT API calls "temperature" - a measure of the amount of randomness used to generate output. Lower temperature leads to less random (more conventional) outputs, and higher temperature leads to more random (more potentially creative) outputs. To avoid confusion, we name this "Creativity" and provide  three choices: Low, Medium, and High, corresponding to temperatures of 0.7, 1.0, and 1.3, respectively.

In each of the above cases, the Ideator system generates one or more ideas for each move. Users can rate each of these ideas with a Thumbs Up or Thumbs Down button, and they can Bookmark ideas they particularly like to save these ideas in their personal collection.

\subsection{Implementation}

\begin{figure}[h!]
  \centering
  \includegraphics[width=\linewidth]{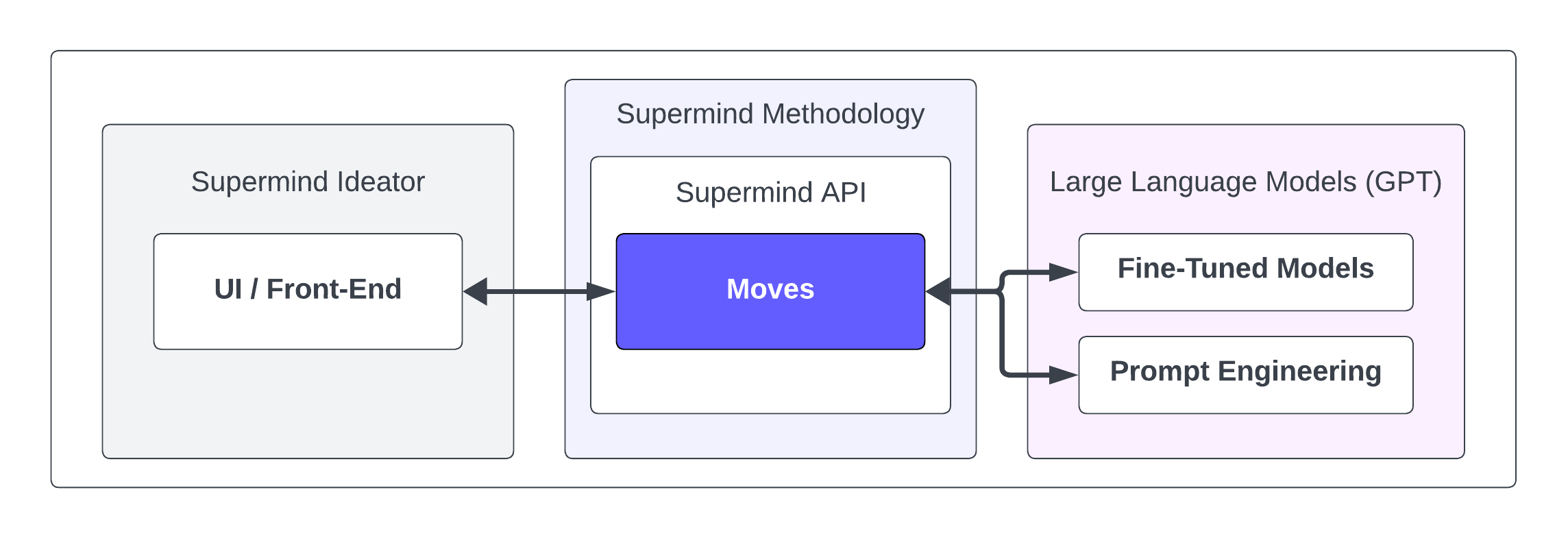}
  \vspace{-20px}
  \caption{The Layers of the Supermind Ideator System. These consist of the User Interface, the API, and the LLM being used. The interface is on top and is based on a React framework. The API is in the middle and connects the interface to the LLM through the software that implements our Moves. The LLM is at the bottom and includes specially fine-tuned versions.}
  \Description{The Layers of the Supermind Ideator System. These consist of the User Interface, the API, and the LLM being used. The interface is on top and is based on a React framework. The API is in the middle and connects the interface to the LLM through the software that implements our Moves. The LLM is at the bottom and includes specially fine-tuned versions.}
  \label{fig:ideatorArch}
\end{figure}

The Ideator system includes multiple layers that work together to provide structure and guidance to an otherwise open-ended problem. These layers (shown in Fig \ref{fig:ideatorArch}) include the \textit{User Interface}, the \textit{API}, and the \textit{LLM}. 

To increase the ease of maintaining and extending the system, we built an API (Application Programming Interface) to separate the client-side user interface logic from server-side processing logic.

We expect this API-based approach to also be valuable in making the Supermind Ideator moves available to others who might want to build upon or extend our methodology beyond the initial web interface we have provided. The API simplifies each move into a simple set of text strings with most of the formatting and interaction with OpenAI's GPT models abstracted away to optional parameters. We elaborate upon how to use the API in the Appendix.

\subsection{Fine-Tuning}

While the general knowledge embedded within GPT's base model can produce interesting and fitting output with relatively limited guidance (zero-shot and few-shot prompts), it seemed apt to augment this base knowledge with specialized information so that more precise and useful ideas could also be generated. To accomplish this, we curated approximately 1600 examples of real-world organizational practices that solved problems using various innovative approaches. We organized this case study corpus with input labels that connected the problem and case study with an associated Cognify or Groupify move.

This enabled us to fine-tune GPT so the new model could map between input language, output prediction, and Supermind move. This new fine-tuned model is much less constrained than our zero-shot and few-shot moves. While these earlier moves answer a question or problem with another question or idea based on the base GPT model and our prompting, the fine-tuned approach changes the GPT prediction model so that output more tightly reflects the input training data. As our training data was a corpus of case studies, the fine-tuned model produces output more akin to those case studies. This produces an interesting challenge and opportunity as the GPT system tends to produce fictional content in a coherent and sensible manner, reading as if it is stating facts.

\vspace{10px}

For example, when prompted with the problem: "I want to improve the way companies retrain employees whose jobs have been replaced by automation," we get the following examples:

\small
\begin{itemize}
    \item Groupify - Market: "Swiss logistics company Travida uses a collective intelligence method to help with employee retraining in the face of automation replacement. Its platform, called Morpheus, allows for crowdsourcing of training needs, as well as for matching of people with training openings"
    \item Cognify - Remember: "Microsoft has been experimenting with virtual retraining environments that mix real and virtual reality to allow workers to experience new scenarios. The idea is that learning should be less dependent on physical location, and could be reimagined as a continuous, fluid process that doesn't end with graduation."
\end{itemize}
\normalsize

Even though both of these examples sound factual, they are actually "hallucinations," fictional combinations of information from different case studies in our corpus, the base GPT model, and other predictions made by the transformer's neural network. 

If our goal were to help users retrieve factual content, this would be a problem. But as noted above, since our goal is to help users come up with creative ideas, we think these fictitious hallucinations are more likely to be a feature than a bug. Creativity researchers have found, for example, that exposure to other creative ideas can often be useful triggers for stimulating people to have creative ideas they would never have had otherwise \cite{fink_stimulating_2012}. To avoid misconceptions on the part of our users, however, we label these outputs as "possible (maybe fictitious) idea(s)."

\section{Evaluation}

In order to evaluate the system, we ran a formative study with 40 participants. The participants were professional consultants, designers, and others who work in and around organizational innovation practices in a variety of industries, including management consulting, IT services, pharmaceuticals, and universities. Most participants had at least 10 to 20 years of professional experience in their particular field. 

All study activity occurred remotely through online meetings which averaged about 45 minutes in length. The researcher ran a think-aloud protocol with each participant, first explaining the Supermind Ideator application, and then asking the participant to suggest a problem statement to kick off the session. As the system generated ideas, participants were asked to reflect on and respond to the output. As this was a formative study, we collected (a) UX insights related to which aspects of the application 'made sense' to participants and which did not, and (b) measures of professional applicability, such as where and how participants might make use of the application in their professional lives.

Generally, participants responded positively to the application, expressing excitement at the potential for the Supermind Ideator to help them innovate. As one participant put it, "the tool could be invaluable to help us, and our clients, explore multiple dimensions of a problem at the inception of a project. Today we struggle to ensure that all projects and all teams think broadly and systematically about the contours of the problem space." Another noted that "the [Supermind Ideator] helps effortlessly bring up questions that my colleagues and I need to have an answer to." A third remarked that the system "can help me and especially my teams break down client problems when we don't have an intimate understanding of their domain space. It helps us come up to speed much, much faster." Yet another noted "it helps us look at many different angles, and recombine them to get to new ideas. And it makes that discovery so much faster."

One consultant summarized the potential use of the tool by individuals, not teams, as follows:
\small
\begin{quote}
    "I have been an innovation 'army of one' on many projects and I find this tool very helpful because it helps you scale very fast into obtaining the results that you would've gotten out of a small workshop, [that] could possibly take you a day to organize; instead you managed to do this in no more than five minutes."
\end{quote}
\normalsize

Some participants also brought to light limitations of the current system that suggest potential ways it could be extended in the future. One participant noted, "[the Supermind Ideator] could be much more powerful if it could explore parts of the problems farther away from where I pointed it at... the tool kept thinking about how to make the [problem] more efficient while the real answer is to look at the upstream possibility... It could have abstracted one level more, or looked at up and downstream problems, to get there." Another remarked that "the tool structure is very portable to a variety of problems, and can be easily repurposed."

Finally, it was noteworthy that our participants organically realized the power of augmenting human abilities rather than replacing them, noting "it is fantastic to see that the tool is intended to help people, instead of sidelining them." 

\section{Discussion}

As generative systems (and LLMs in particular) continue to be developed and deployed into real world applications, a few key topics remain critical for awareness not only for Ideator but for the research community at large.

\subsection{Specialized front ends}
While chat-based LLMs are usually very easy to use and can be used for a wide range of applications, this work demonstrates that there are also situations where specialized front ends can be valuable. Most users are not expert prompt engineers, and they may not be very good at using a chat-based interface to carry out a structured process for a specific kind of task. But as the Supermind Ideator illustrates for the task of creative problem-solving, a specialized front end can include (a) carefully selected prompts that are known to achieve good results for this task and (b) customized UIs that greatly simplify using these prompts in a structured process with parameters specific to a user’s problem.

\subsection{Systematic processes}
For tasks with codified methodologies that humans use, there is value in guiding users through a process. Unlike a chat-based interface that tries to gratify the user at each step of the chat, a specialized front-end can act more like a human facilitator who guides the user through the activities specified in the methodology. For example, in the creative problem-solving domain, human facilitators often take participants through a long process to break them out of their own fixations and find new ways of understanding the problem before exploring possible solutions. The Supermind Ideator supports this process without human facilitators by guiding users to explore the problem space first and then the solution space. 

\subsection{Recombination of ideas}
An important part of creative problem-solving is often the organization of fragments of ideas that can be recombined with others. Our UI supports simplified collection of highlights by letting users bookmark key results, reducing the potential for good ideas to get lost in the avalanche of text that LLMs can produce. Interfaces should be designed in ways that explicitly support idea synthesis and recombination, for example rating the inner elements of an idea (at the word/sentence level) rather than simply rating an entire idea.

\subsection{Community-based extension}
While we have not yet demonstrated this capability, the structure of the Ideator suggests how a specialized front end and flexible back end can help create a platform for community creation and evolution. Collective software development by an open-source community, for example, can improve or add individual “moves” or run combinations of moves within custom UIs. We hope that our contribution scaffolds future design and development of tools that augment the collective ability of many others.

\section{Future work}

It appears that the Ideator has the potential to provide substantial assistance to humans doing creative problem-solving. Much work remains to be done for this potential to be fully realized, and more complete, quantitative evaluations of factors such as the user experience of people using the system and the speed and quality of the ideas they generate are currently underway. We also see at least three short-term paths forward to improve Ideator:

\subsection{Adding and Improving Moves}
One of our short-term goals is to refine the prompts and fine-tuning of our moves to improve the quality of the ideas the moves generate. In addition, it seems very promising (and feasible) to add support for new types of moves. Some of these moves might be relevant for creative problem-solving in any domain. For example, moves might be developed for techniques such as lateral thinking \cite{de1970lateral} or "5 W's and 1 H" \cite{brown_2009}.

It would also be possible to include moves to support other methodologies for thinking about business questions, such as Porter's 5-forces \cite{porter_1998} or Blue Ocean Strategy \cite{kim_mauborgne_renee_2017}. And it would potentially be possible to add moves to support other specific topic domains such as mechanical design or building architecture. 

We hope that by having an extensible and open-source API for the Supermind Ideator, communities of professionals, researchers, consultants, and others could contribute to a growing collective of knowledge about how to guide LLMs to help with specific types of ideation and creative problem solving.

\subsection{Evaluating Ideas}

Currently, the moves we have implemented only cover the divergent aspects of idea generation: generating possible ways of re-framing the problem and generating possible solutions to the problem. As the double diamond process suggests, however, evaluating and selecting among these possibilities is also necessary to be able to actually use the results.

For instance, as described above, the thumbs-up, thumbs-down, and bookmark buttons provide simple tools for evaluating ideas. With these features, users can quickly skim through their previous ideas and find things they liked, disliked, or saved. 

Looking forward we feel that idea evaluation can be greatly expanded upon in a number of ways such as sharing ratings in a group and using previous ratings to recommend items in a new situation.

\subsection{Adding Exploration Modes}

The Supermind Ideator currently produces output in a single thread of ideas. Each round of idea generation comes after the next in a linear manner. Re-running moves allows for a nesting of output, but that is as far as we currently go. From our user evaluation we found that people do not always work in linear patterns of thought and would occasionally find themselves wanting to combine outputs from different moves or otherwise reorganize output in a manner that allowed the user to see clusters of thought and reflect on them in non-threaded and non-linear manners. 

Taking into consideration the power of affinity diagrams when brainstorming, we believe a more flexible and dynamic interface could be designed to enable users to connect their input, output, and associated ideas in less restrictive ways. For example, by giving 'sticky note' like flexibility to the interface there might be increased capacity for synthesis of output. 

Furthermore, if the idea generation process is thought of as traversing a graph, then an interface that treats each idea as a node could allow for one to see the spaces they have already deeply explored vs. those they have not yet tapped into and either help broaden exploration or narrow focus. 

\section{Conclusion}

In this paper, we have suggested how large language models can be used to help people do creative problem-solving in a very broad range of areas. We have also shown how these models can be further customized to help with creative problem-solving in a specific topic area--in this case, the area of designing groups of people and/or computers. We believe that this approach can also be extended to include many other general-purpose creativity techniques and many other specialized techniques for use in this and other topic areas.

In the long run, we hope that tools like this will help superminds--composed of people and computers working together--deal more creatively with our most important problems in business, government, science, and many other areas of society.

\bibliographystyle{ACM-Reference-Format}
\bibliography{ideator_refs}


\begin{thebibliography}{52}


\ifx \showCODEN    \undefined \def \showCODEN     #1{\unskip}     \fi
\ifx \showDOI      \undefined \def \showDOI       #1{#1}\fi
\ifx \showISBNx    \undefined \def \showISBNx     #1{\unskip}     \fi
\ifx \showISBNxiii \undefined \def \showISBNxiii  #1{\unskip}     \fi
\ifx \showISSN     \undefined \def \showISSN      #1{\unskip}     \fi
\ifx \showLCCN     \undefined \def \showLCCN      #1{\unskip}     \fi
\ifx \shownote     \undefined \def \shownote      #1{#1}          \fi
\ifx \showarticletitle \undefined \def \showarticletitle #1{#1}   \fi
\ifx \showURL      \undefined \def \showURL       {\relax}        \fi
\providecommand\bibfield[2]{#2}
\providecommand\bibinfo[2]{#2}
\providecommand\natexlab[1]{#1}
\providecommand\showeprint[2][]{arXiv:#2}

\bibitem[dsc({[n.\,d.]})]%
        {dschool_design_bootleg}
 \bibinfo{year}{[n.\,d.]}\natexlab{}.
\newblock \bibinfo{title}{Design {Thinking} {Bootleg}}.
\newblock
\newblock
\urldef\tempurl%
\url{https://dschool.stanford.edu/resources/design-thinking-bootleg}
\showURL{%
\tempurl}


\bibitem[noa(2005)]%
        {noauthor_framework_2005}
 \bibinfo{year}{2005}\natexlab{}.
\newblock \bibinfo{title}{Framework for {Innovation}: {Design} {Council}'s evolved {Double} {Diamond}}.
\newblock
\newblock
\urldef\tempurl%
\url{https://www.designcouncil.org.uk/our-work/skills-learning/tools-frameworks/framework-for-innovation-design-councils-evolved-double-diamond/}
\showURL{%
\tempurl}


\bibitem[Bender et~al\mbox{.}(2021)]%
        {bender_dangers_2021}
\bibfield{author}{\bibinfo{person}{Emily~M. Bender}, \bibinfo{person}{Timnit Gebru}, \bibinfo{person}{Angelina McMillan-Major}, {and} \bibinfo{person}{Shmargaret Shmitchell}.} \bibinfo{year}{2021}\natexlab{}.
\newblock \showarticletitle{On the {Dangers} of {Stochastic} {Parrots}: {Can} {Language} {Models} {Be} {Too} {Big}}. In \bibinfo{booktitle}{\emph{Proceedings of the 2021 {ACM} {Conference} on {Fairness}, {Accountability}, and {Transparency}}} \emph{(\bibinfo{series}{{FAccT} '21})}. \bibinfo{publisher}{Association for Computing Machinery}, \bibinfo{address}{New York, NY, USA}, \bibinfo{pages}{610--623}.
\newblock
\showISBNx{978-1-4503-8309-7}
\urldef\tempurl%
\url{https://doi.org/10.1145/3442188.3445922}
\showDOI{\tempurl}


\bibitem[Brown(2009)]%
        {brown_2009}
\bibfield{author}{\bibinfo{person}{Tom Brown}.} \bibinfo{year}{2009}\natexlab{}.
\newblock \bibinfo{booktitle}{\emph{Change by design}}.
\newblock \bibinfo{publisher}{HARPER COLLINS}.
\newblock


\bibitem[Brown et~al\mbox{.}(2020)]%
        {brown_language_2020}
\bibfield{author}{\bibinfo{person}{Tom Brown}, \bibinfo{person}{Benjamin Mann}, \bibinfo{person}{Nick Ryder}, \bibinfo{person}{Melanie Subbiah}, \bibinfo{person}{Jared~D Kaplan}, \bibinfo{person}{Prafulla Dhariwal}, \bibinfo{person}{Arvind Neelakantan}, \bibinfo{person}{Pranav Shyam}, \bibinfo{person}{Girish Sastry}, \bibinfo{person}{Amanda Askell}, \bibinfo{person}{Sandhini Agarwal}, \bibinfo{person}{Ariel Herbert-Voss}, \bibinfo{person}{Gretchen Krueger}, \bibinfo{person}{Tom Henighan}, \bibinfo{person}{Rewon Child}, \bibinfo{person}{Aditya Ramesh}, \bibinfo{person}{Daniel Ziegler}, \bibinfo{person}{Jeffrey Wu}, \bibinfo{person}{Clemens Winter}, \bibinfo{person}{Chris Hesse}, \bibinfo{person}{Mark Chen}, \bibinfo{person}{Eric Sigler}, \bibinfo{person}{Mateusz Litwin}, \bibinfo{person}{Scott Gray}, \bibinfo{person}{Benjamin Chess}, \bibinfo{person}{Jack Clark}, \bibinfo{person}{Christopher Berner}, \bibinfo{person}{Sam McCandlish}, \bibinfo{person}{Alec Radford}, \bibinfo{person}{Ilya Sutskever}, {and}
  \bibinfo{person}{Dario Amodei}.} \bibinfo{year}{2020}\natexlab{}.
\newblock \showarticletitle{Language {Models} are {Few}-{Shot} {Learners}}. In \bibinfo{booktitle}{\emph{Advances in {Neural} {Information} {Processing} {Systems}}}, Vol.~\bibinfo{volume}{33}. \bibinfo{publisher}{Curran Associates, Inc.}, \bibinfo{pages}{1877--1901}.
\newblock
\urldef\tempurl%
\url{https://proceedings.neurips.cc/paper/2020/hash/1457c0d6bfcb4967418bfb8ac142f64a-Abstract.html}
\showURL{%
\tempurl}


\bibitem[Bygstad(2010)]%
        {bygstad_generative_2010}
\bibfield{author}{\bibinfo{person}{Bendik Bygstad}.} \bibinfo{year}{2010}\natexlab{}.
\newblock \showarticletitle{Generative mechanisms for innovation in information infrastructures}.
\newblock \bibinfo{journal}{\emph{Information and Organization}} \bibinfo{volume}{20}, \bibinfo{number}{3} (\bibinfo{date}{July} \bibinfo{year}{2010}), \bibinfo{pages}{156--168}.
\newblock
\showISSN{1471-7727}
\urldef\tempurl%
\url{https://doi.org/10.1016/j.infoandorg.2010.07.001}
\showDOI{\tempurl}


\bibitem[Cardoso and Badke-Schaub(2011)]%
        {cardoso_fixation_2011}
\bibfield{author}{\bibinfo{person}{Carlos Cardoso} {and} \bibinfo{person}{Petra Badke-Schaub}.} \bibinfo{year}{2011}\natexlab{}.
\newblock \showarticletitle{Fixation or {Inspiration}: {Creative} {Problem} {Solving} in {Design}}.
\newblock \bibinfo{journal}{\emph{The Journal of Creative Behavior}} \bibinfo{volume}{45}, \bibinfo{number}{2} (\bibinfo{date}{June} \bibinfo{year}{2011}), \bibinfo{pages}{77--82}.
\newblock
\showISSN{00220175}
\urldef\tempurl%
\url{https://doi.org/10.1002/j.2162-6057.2011.tb01086.x}
\showDOI{\tempurl}


\bibitem[Chen and Terken(2023)]%
        {chen_design_2023}
\bibfield{author}{\bibinfo{person}{Fang Chen} {and} \bibinfo{person}{Jacques Terken}.} \bibinfo{year}{2023}\natexlab{}.
\newblock \showarticletitle{Design {Process}}.
\newblock In \bibinfo{booktitle}{\emph{Automotive {Interaction} {Design}: {From} {Theory} to {Practice}}}, \bibfield{editor}{\bibinfo{person}{Fang Chen} {and} \bibinfo{person}{Jacques Terken}} (Eds.). \bibinfo{publisher}{Springer Nature}, \bibinfo{address}{Singapore}, \bibinfo{pages}{165--179}.
\newblock
\showISBNx{978-981-19344-8-3}
\urldef\tempurl%
\url{https://doi.org/10.1007/978-981-19-3448-3_10}
\showDOI{\tempurl}


\bibitem[Cross(1982)]%
        {cross_designerly_1982}
\bibfield{author}{\bibinfo{person}{Nigel Cross}.} \bibinfo{year}{1982}\natexlab{}.
\newblock \showarticletitle{Designerly ways of knowing}.
\newblock \bibinfo{journal}{\emph{Design Studies}} \bibinfo{volume}{3}, \bibinfo{number}{4} (\bibinfo{date}{Oct.} \bibinfo{year}{1982}), \bibinfo{pages}{221--227}.
\newblock
\showISSN{0142-694X}
\urldef\tempurl%
\url{https://doi.org/10.1016/0142-694X(82)90040-0}
\showDOI{\tempurl}


\bibitem[Dam and Siang(2020)]%
        {dam_design_2020}
\bibfield{author}{\bibinfo{person}{Rikke~Friis Dam} {and} \bibinfo{person}{Teo~Yu Siang}.} \bibinfo{year}{2020}\natexlab{}.
\newblock \showarticletitle{Design thinking: a quick overview}.
\newblock  (\bibinfo{date}{June} \bibinfo{year}{2020}).
\newblock
\urldef\tempurl%
\url{https://apo.org.au/node/306478}
\showURL{%
\tempurl}
\newblock
\shownote{Publisher: Interaction Design Foundation}.


\bibitem[De~Bono(1970)]%
        {de1970lateral}
\bibfield{author}{\bibinfo{person}{Edward De~Bono}.} \bibinfo{year}{1970}\natexlab{}.
\newblock \showarticletitle{Lateral thinking}.
\newblock \bibinfo{journal}{\emph{New York}} (\bibinfo{year}{1970}).
\newblock


\bibitem[Dorst and Cross(2001)]%
        {dorst_creativity_2001}
\bibfield{author}{\bibinfo{person}{Kees Dorst} {and} \bibinfo{person}{Nigel Cross}.} \bibinfo{year}{2001}\natexlab{}.
\newblock \showarticletitle{Creativity in the design process: co-evolution of problem–solution}.
\newblock \bibinfo{journal}{\emph{Design Studies}} \bibinfo{volume}{22}, \bibinfo{number}{5} (\bibinfo{date}{Sept.} \bibinfo{year}{2001}), \bibinfo{pages}{425--437}.
\newblock
\showISSN{0142-694X}
\urldef\tempurl%
\url{https://doi.org/10.1016/S0142-694X(01)00009-6}
\showDOI{\tempurl}


\bibitem[Eckert and Stacey(2000)]%
        {eckert_sources_2000}
\bibfield{author}{\bibinfo{person}{Claudia Eckert} {and} \bibinfo{person}{Martin Stacey}.} \bibinfo{year}{2000}\natexlab{}.
\newblock \showarticletitle{Sources of inspiration: a language of design}.
\newblock \bibinfo{journal}{\emph{Design Studies}} \bibinfo{volume}{21}, \bibinfo{number}{5} (\bibinfo{date}{Sept.} \bibinfo{year}{2000}), \bibinfo{pages}{523--538}.
\newblock
\showISSN{0142-694X}
\urldef\tempurl%
\url{https://doi.org/10.1016/S0142-694X(00)00022-3}
\showDOI{\tempurl}


\bibitem[Fink et~al\mbox{.}(2012)]%
        {fink_stimulating_2012}
\bibfield{author}{\bibinfo{person}{Andreas Fink}, \bibinfo{person}{Karl Koschutnig}, \bibinfo{person}{Mathias Benedek}, \bibinfo{person}{Gernot Reishofer}, \bibinfo{person}{Anja Ischebeck}, \bibinfo{person}{Elisabeth~M. Weiss}, {and} \bibinfo{person}{Franz Ebner}.} \bibinfo{year}{2012}\natexlab{}.
\newblock \showarticletitle{Stimulating creativity via the exposure to other people's ideas}.
\newblock \bibinfo{journal}{\emph{Human Brain Mapping}} \bibinfo{volume}{33}, \bibinfo{number}{11} (\bibinfo{year}{2012}), \bibinfo{pages}{2603--2610}.
\newblock
\showISSN{1097-0193}
\urldef\tempurl%
\url{https://doi.org/10.1002/hbm.21387}
\showDOI{\tempurl}
\newblock
\shownote{\_eprint: https://onlinelibrary.wiley.com/doi/pdf/10.1002/hbm.21387}.


\bibitem[for Collective~Intelligence({[n.\,d.]})]%
        {noauthor_supermind_nodate}
\bibfield{author}{\bibinfo{person}{MIT~Center for Collective~Intelligence}.} \bibinfo{year}{[n.\,d.]}\natexlab{}.
\newblock \bibinfo{title}{Supermind {Design} {Primer} {\textbar} {MIT} {Center} for {Collective} {Intelligence}}.
\newblock
\newblock
\urldef\tempurl%
\url{https://cci.mit.edu/supermind-design-primer/}
\showURL{%
\tempurl}


\bibitem[Frich et~al\mbox{.}(2018)]%
        {frich_twenty_2018}
\bibfield{author}{\bibinfo{person}{Jonas Frich}, \bibinfo{person}{Michael Mose~Biskjaer}, {and} \bibinfo{person}{Peter Dalsgaard}.} \bibinfo{year}{2018}\natexlab{}.
\newblock \showarticletitle{Twenty {Years} of {Creativity} {Research} in {Human}-{Computer} {Interaction}: {Current} {State} and {Future} {Directions}}. In \bibinfo{booktitle}{\emph{Proceedings of the 2018 {Designing} {Interactive} {Systems} {Conference}}} \emph{(\bibinfo{series}{{DIS} '18})}. \bibinfo{publisher}{Association for Computing Machinery}, \bibinfo{address}{New York, NY, USA}, \bibinfo{pages}{1235--1257}.
\newblock
\showISBNx{978-1-4503-5198-0}
\urldef\tempurl%
\url{https://doi.org/10.1145/3196709.3196732}
\showDOI{\tempurl}


\bibitem[Gabora(2002)]%
        {gabora_cognitive_2002}
\bibfield{author}{\bibinfo{person}{Liane Gabora}.} \bibinfo{year}{2002}\natexlab{}.
\newblock \showarticletitle{Cognitive mechanisms underlying the creative process}. In \bibinfo{booktitle}{\emph{Proceedings of the 4th conference on {Creativity} \& cognition}} \emph{(\bibinfo{series}{C\&amp;{C} '02})}. \bibinfo{publisher}{Association for Computing Machinery}, \bibinfo{address}{New York, NY, USA}, \bibinfo{pages}{126--133}.
\newblock
\showISBNx{978-1-58113-465-0}
\urldef\tempurl%
\url{https://doi.org/10.1145/581710.581730}
\showDOI{\tempurl}


\bibitem[Gilhooly et~al\mbox{.}(2007)]%
        {gilhooly_divergent_2007}
\bibfield{author}{\bibinfo{person}{K.~J. Gilhooly}, \bibinfo{person}{E. Fioratou}, \bibinfo{person}{S.~H. Anthony}, {and} \bibinfo{person}{V. Wynn}.} \bibinfo{year}{2007}\natexlab{}.
\newblock \showarticletitle{Divergent thinking: {Strategies} and executive involvement in generating novel uses for familiar objects}.
\newblock \bibinfo{journal}{\emph{British Journal of Psychology}} \bibinfo{volume}{98}, \bibinfo{number}{4} (\bibinfo{year}{2007}), \bibinfo{pages}{611--625}.
\newblock
\showISSN{2044-8295}
\urldef\tempurl%
\url{https://doi.org/10.1111/j.2044-8295.2007.tb00467.x}
\showDOI{\tempurl}
\newblock
\shownote{\_eprint: https://onlinelibrary.wiley.com/doi/pdf/10.1111/j.2044-8295.2007.tb00467.x}.


\bibitem[Goodman et~al\mbox{.}(2022)]%
        {goodman_lampost_2022}
\bibfield{author}{\bibinfo{person}{Steven~M. Goodman}, \bibinfo{person}{Erin Buehler}, \bibinfo{person}{Patrick Clary}, \bibinfo{person}{Andy Coenen}, \bibinfo{person}{Aaron Donsbach}, \bibinfo{person}{Tiffanie~N. Horne}, \bibinfo{person}{Michal Lahav}, \bibinfo{person}{Robert MacDonald}, \bibinfo{person}{Rain~Breaw Michaels}, \bibinfo{person}{Ajit Narayanan}, \bibinfo{person}{Mahima Pushkarna}, \bibinfo{person}{Joel Riley}, \bibinfo{person}{Alex Santana}, \bibinfo{person}{Lei Shi}, \bibinfo{person}{Rachel Sweeney}, \bibinfo{person}{Phil Weaver}, \bibinfo{person}{Ann Yuan}, {and} \bibinfo{person}{Meredith~Ringel Morris}.} \bibinfo{year}{2022}\natexlab{}.
\newblock \showarticletitle{{LaMPost}: {Design} and {Evaluation} of an {AI}-assisted {Email} {Writing} {Prototype} for {Adults} with {Dyslexia}}. In \bibinfo{booktitle}{\emph{Proceedings of the 24th {International} {ACM} {SIGACCESS} {Conference} on {Computers} and {Accessibility}}} \emph{(\bibinfo{series}{{ASSETS} '22})}. \bibinfo{publisher}{Association for Computing Machinery}, \bibinfo{address}{New York, NY, USA}, \bibinfo{pages}{1--18}.
\newblock
\showISBNx{978-1-4503-9258-7}
\urldef\tempurl%
\url{https://doi.org/10.1145/3517428.3544819}
\showDOI{\tempurl}


\bibitem[Gregg(2010)]%
        {gregg_designing_2010}
\bibfield{author}{\bibinfo{person}{Dawn~G. Gregg}.} \bibinfo{year}{2010}\natexlab{}.
\newblock \showarticletitle{Designing for collective intelligence}.
\newblock \bibinfo{journal}{\emph{Commun. ACM}} \bibinfo{volume}{53}, \bibinfo{number}{4} (\bibinfo{date}{April} \bibinfo{year}{2010}), \bibinfo{pages}{134--138}.
\newblock
\showISSN{0001-0782}
\urldef\tempurl%
\url{https://doi.org/10.1145/1721654.1721691}
\showDOI{\tempurl}


\bibitem[Griebel et~al\mbox{.}(2020)]%
        {griebel_augmented_2020}
\bibfield{author}{\bibinfo{person}{Matthias Griebel}, \bibinfo{person}{Christoph Flath}, {and} \bibinfo{person}{Sascha Friesike}.} \bibinfo{year}{2020}\natexlab{}.
\newblock \showarticletitle{{AUGMENTED} {CREATIVITY}: {LEVERAGING} {ARTIFICIAL} {INTELLIGENCE} {FOR} {IDEA} {GENERATION} {IN} {THE} {CREATIVE} {SPHERE}}.
\newblock  (\bibinfo{year}{2020}).
\newblock


\bibitem[Howard et~al\mbox{.}(2010)]%
        {howard_use_2010}
\bibfield{author}{\bibinfo{person}{T.~J. Howard}, \bibinfo{person}{E.~A. Dekoninck}, {and} \bibinfo{person}{S.~J. Culley}.} \bibinfo{year}{2010}\natexlab{}.
\newblock \showarticletitle{The use of creative stimuli at early stages of industrial product innovation}.
\newblock \bibinfo{journal}{\emph{Research in Engineering Design}} \bibinfo{volume}{21}, \bibinfo{number}{4} (\bibinfo{date}{Oct.} \bibinfo{year}{2010}), \bibinfo{pages}{263--274}.
\newblock
\showISSN{1435-6066}
\urldef\tempurl%
\url{https://doi.org/10.1007/s00163-010-0091-4}
\showDOI{\tempurl}


\bibitem[Jansson and Smith(1991)]%
        {jansson_design_1991}
\bibfield{author}{\bibinfo{person}{David~G. Jansson} {and} \bibinfo{person}{Steven~M. Smith}.} \bibinfo{year}{1991}\natexlab{}.
\newblock \showarticletitle{Design fixation}.
\newblock \bibinfo{journal}{\emph{Design Studies}} \bibinfo{volume}{12}, \bibinfo{number}{1} (\bibinfo{date}{Jan.} \bibinfo{year}{1991}), \bibinfo{pages}{3--11}.
\newblock
\showISSN{0142-694X}
\urldef\tempurl%
\url{https://doi.org/10.1016/0142-694X(91)90003-F}
\showDOI{\tempurl}


\bibitem[Kim and Renee(2017)]%
        {kim_mauborgne_renee_2017}
\bibfield{author}{\bibinfo{person}{W.~Chan Kim} {and} \bibinfo{person}{Mauborgne Renee}.} \bibinfo{year}{2017}\natexlab{}.
\newblock \bibinfo{booktitle}{\emph{Blue Ocean Shift: Beyond competing: Proven steps to inspire confidence and seize new growth}}.
\newblock \bibinfo{publisher}{Hachette Books}.
\newblock


\bibitem[Klein and Convertino(2014)]%
        {klein2014embarrassment}
\bibfield{author}{\bibinfo{person}{Mark Klein} {and} \bibinfo{person}{Gregorio Convertino}.} \bibinfo{year}{2014}\natexlab{}.
\newblock \showarticletitle{An embarrassment of riches}.
\newblock \bibinfo{journal}{\emph{Commun. ACM}} \bibinfo{volume}{57}, \bibinfo{number}{11} (\bibinfo{year}{2014}), \bibinfo{pages}{40--42}.
\newblock


\bibitem[Koberg and Bagnall(1974)]%
        {koberg1974universal}
\bibfield{author}{\bibinfo{person}{Don Koberg} {and} \bibinfo{person}{Jim Bagnall}.} \bibinfo{year}{1974}\natexlab{}.
\newblock \showarticletitle{The Universal Traveler, A Soft-Systems Guide to: Creativity, Problem Solving, and the Process of Reaching Goals.[Revised Edition.].}
\newblock  (\bibinfo{year}{1974}).
\newblock


\bibitem[Koppineni et~al\mbox{.}(2022)]%
        {koppineni2022supermind}
\bibfield{author}{\bibinfo{person}{Akhilesh Koppineni}, \bibinfo{person}{David~Sun Kong}, {and} \bibinfo{person}{Thomas~W Malone}.} \bibinfo{year}{2022}\natexlab{}.
\newblock \showarticletitle{Supermind Design for Responding to Covid-19: A case study of university students generating innovative ideas for a societal problem}.
\newblock  (\bibinfo{year}{2022}).
\newblock


\bibitem[Laubacher et~al\mbox{.}(2020)]%
        {laubacher_using_2020}
\bibfield{author}{\bibinfo{person}{Robert Laubacher}, \bibinfo{person}{Gianni Giacomelli}, \bibinfo{person}{Kathleen Kennedy}, \bibinfo{person}{David~Sun Kong}, \bibinfo{person}{Annalyn Bachmann}, \bibinfo{person}{Katharina Kramer}, \bibinfo{person}{Paul Schlag}, {and} \bibinfo{person}{Thomas~W. Malone}.} \bibinfo{year}{2020}\natexlab{}.
\newblock \bibinfo{title}{Using a {Supermind} to {Design} a {Supermind}: {A} {Case} {Study} of {University} {Researchers} and {Corporate} {Executives} {Co}-{Designing} an {Innovative} {Healthcare} {Concept}}.
\newblock
\newblock
\urldef\tempurl%
\url{https://doi.org/10.2139/ssrn.3601059}
\showDOI{\tempurl}


\bibitem[Lee and Jin(2019)]%
        {lee_how_2019}
\bibfield{author}{\bibinfo{person}{Jung-Yong Lee} {and} \bibinfo{person}{Chang-Hyun Jin}.} \bibinfo{year}{2019}\natexlab{}.
\newblock \showarticletitle{How {Collective} {Intelligence} {Fosters} {Incremental} {Innovation}}.
\newblock \bibinfo{journal}{\emph{Journal of Open Innovation: Technology, Market, and Complexity}} \bibinfo{volume}{5}, \bibinfo{number}{3} (\bibinfo{date}{Sept.} \bibinfo{year}{2019}), \bibinfo{pages}{53}.
\newblock
\showISSN{2199-8531}
\urldef\tempurl%
\url{https://doi.org/10.3390/joitmc5030053}
\showDOI{\tempurl}


\bibitem[Lee et~al\mbox{.}(2023)]%
        {lee_prompts_2023}
\bibfield{author}{\bibinfo{person}{Terence Lee}, \bibinfo{person}{Lauren O’Mahony}, {and} \bibinfo{person}{Pia Lebeck}.} \bibinfo{year}{2023}\natexlab{}.
\newblock \showarticletitle{Prompts for {Creativity}}.
\newblock In \bibinfo{booktitle}{\emph{Creativity and {Innovation}: {Everyday} {Dynamics} and {Practice}}}, \bibfield{editor}{\bibinfo{person}{Terence Lee}, \bibinfo{person}{Lauren O'Mahony}, {and} \bibinfo{person}{Pia Lebeck}} (Eds.). \bibinfo{publisher}{Springer Nature}, \bibinfo{address}{Singapore}, \bibinfo{pages}{85--115}.
\newblock
\showISBNx{978-981-19888-0-6}
\urldef\tempurl%
\url{https://doi.org/10.1007/978-981-19-8880-6_4}
\showDOI{\tempurl}


\bibitem[Linsey et~al\mbox{.}(2010)]%
        {linsey_study_2010}
\bibfield{author}{\bibinfo{person}{J.~S. Linsey}, \bibinfo{person}{I. Tseng}, \bibinfo{person}{K. Fu}, \bibinfo{person}{J. Cagan}, \bibinfo{person}{K.~L. Wood}, {and} \bibinfo{person}{C. Schunn}.} \bibinfo{year}{2010}\natexlab{}.
\newblock \showarticletitle{A {Study} of {Design} {Fixation}, {Its} {Mitigation} and {Perception} in {Engineering} {Design} {Faculty}}.
\newblock \bibinfo{journal}{\emph{Journal of Mechanical Design}} \bibinfo{volume}{132}, \bibinfo{number}{4} (\bibinfo{date}{April} \bibinfo{year}{2010}), \bibinfo{pages}{041003}.
\newblock
\showISSN{1050-0472, 1528-9001}
\urldef\tempurl%
\url{https://doi.org/10.1115/1.4001110}
\showDOI{\tempurl}


\bibitem[Lubart(2001)]%
        {lubart_models_2001}
\bibfield{author}{\bibinfo{person}{Todd~I. Lubart}.} \bibinfo{year}{2001}\natexlab{}.
\newblock \showarticletitle{Models of the {Creative} {Process}: {Past}, {Present} and {Future}}.
\newblock \bibinfo{journal}{\emph{Creativity Research Journal}} \bibinfo{volume}{13}, \bibinfo{number}{3-4} (\bibinfo{date}{Oct.} \bibinfo{year}{2001}), \bibinfo{pages}{295--308}.
\newblock
\showISSN{1040-0419}
\urldef\tempurl%
\url{https://doi.org/10.1207/S15326934CRJ1334_07}
\showDOI{\tempurl}
\newblock
\shownote{Publisher: Routledge \_eprint: https://doi.org/10.1207/S15326934CRJ1334\_07}.


\bibitem[Majchrzak and Malhotra(2013)]%
        {majchrzak_towards_2013}
\bibfield{author}{\bibinfo{person}{A. Majchrzak} {and} \bibinfo{person}{A. Malhotra}.} \bibinfo{year}{2013}\natexlab{}.
\newblock \showarticletitle{Towards an information systems perspective and research agenda on crowdsourcing for innovation}.
\newblock \bibinfo{journal}{\emph{The Journal of Strategic Information Systems}} \bibinfo{volume}{22}, \bibinfo{number}{4} (\bibinfo{date}{Dec.} \bibinfo{year}{2013}), \bibinfo{pages}{257--268}.
\newblock
\showISSN{0963-8687}
\urldef\tempurl%
\url{https://doi.org/10.1016/j.jsis.2013.07.004}
\showDOI{\tempurl}


\bibitem[Malone(2018)]%
        {malone2018superminds}
\bibfield{author}{\bibinfo{person}{Thomas~W Malone}.} \bibinfo{year}{2018}\natexlab{}.
\newblock \bibinfo{booktitle}{\emph{Superminds: The surprising power of people and computers thinking together}}.
\newblock \bibinfo{publisher}{Little, Brown Spark}.
\newblock


\bibitem[Malone et~al\mbox{.}(2003)]%
        {malone2003organizing}
\bibfield{author}{\bibinfo{person}{Thomas~W Malone}, \bibinfo{person}{Kevin Crowston}, {and} \bibinfo{person}{George~Arthur Herman}.} \bibinfo{year}{2003}\natexlab{}.
\newblock \bibinfo{booktitle}{\emph{Organizing business knowledge: The MIT process handbook}}.
\newblock \bibinfo{publisher}{MIT press}.
\newblock


\bibitem[Mednick(1963)]%
        {mednick_associative_1963}
\bibfield{author}{\bibinfo{person}{Sarnoff Mednick}.} \bibinfo{year}{1963}\natexlab{}.
\newblock \showarticletitle{The associative basis of the creative process.}
\newblock \bibinfo{journal}{\emph{Psychological Review}} \bibinfo{volume}{69}, \bibinfo{number}{3} (\bibinfo{date}{April} \bibinfo{year}{1963}), \bibinfo{pages}{220}.
\newblock
\showISSN{1939-1471}
\urldef\tempurl%
\url{https://doi.org/10.1037/h0048850}
\showDOI{\tempurl}
\newblock
\shownote{Publisher: US: American Psychological Association}.


\bibitem[Nakata and Hwang(2020)]%
        {nakata_design_2020}
\bibfield{author}{\bibinfo{person}{Cheryl Nakata} {and} \bibinfo{person}{Jiyoung Hwang}.} \bibinfo{year}{2020}\natexlab{}.
\newblock \showarticletitle{Design thinking for innovation: {Composition}, consequence, and contingency}.
\newblock \bibinfo{journal}{\emph{Journal of Business Research}}  \bibinfo{volume}{118} (\bibinfo{date}{Sept.} \bibinfo{year}{2020}), \bibinfo{pages}{117--128}.
\newblock
\showISSN{0148-2963}
\urldef\tempurl%
\url{https://doi.org/10.1016/j.jbusres.2020.06.038}
\showDOI{\tempurl}


\bibitem[Osborn(2012)]%
        {osborn2012applied}
\bibfield{author}{\bibinfo{person}{Alex Osborn}.} \bibinfo{year}{2012}\natexlab{}.
\newblock \bibinfo{booktitle}{\emph{Applied imagination-principles and procedures of creative writing}}.
\newblock \bibinfo{publisher}{Read Books Ltd}.
\newblock


\bibitem[Paulus et~al\mbox{.}(2011)]%
        {paulus2011effects}
\bibfield{author}{\bibinfo{person}{Paul~B Paulus}, \bibinfo{person}{Nicholas~W Kohn}, {and} \bibinfo{person}{Lauren~E Arditti}.} \bibinfo{year}{2011}\natexlab{}.
\newblock \showarticletitle{Effects of quantity and quality instructions on brainstorming}.
\newblock \bibinfo{journal}{\emph{The Journal of Creative Behavior}} \bibinfo{volume}{45}, \bibinfo{number}{1} (\bibinfo{year}{2011}), \bibinfo{pages}{38--46}.
\newblock


\bibitem[Pavie and Carthy(2015)]%
        {pavie_leveraging_2015}
\bibfield{author}{\bibinfo{person}{Xavier Pavie} {and} \bibinfo{person}{Daphne Carthy}.} \bibinfo{year}{2015}\natexlab{}.
\newblock \showarticletitle{Leveraging {Uncertainty}: {A} {Practical} {Approach} to the {Integration} of {Responsible} {Innovation} through {Design} {Thinking}}.
\newblock \bibinfo{journal}{\emph{Procedia - Social and Behavioral Sciences}}  \bibinfo{volume}{213} (\bibinfo{date}{Dec.} \bibinfo{year}{2015}), \bibinfo{pages}{1040--1049}.
\newblock
\showISSN{1877-0428}
\urldef\tempurl%
\url{https://doi.org/10.1016/j.sbspro.2015.11.523}
\showDOI{\tempurl}


\bibitem[Porter(1998)]%
        {porter_1998}
\bibfield{author}{\bibinfo{person}{Michael~E. Porter}.} \bibinfo{year}{1998}\natexlab{}.
\newblock \bibinfo{booktitle}{\emph{Competitive strategy: Techniques for analyzing industries and competitors: With a new introduction}}.
\newblock \bibinfo{publisher}{Free Press}.
\newblock


\bibitem[Robertson and Radcliffe(2008)]%
        {robertson_role_2008}
\bibfield{author}{\bibinfo{person}{Brett Robertson} {and} \bibinfo{person}{David Radcliffe}.} \bibinfo{year}{2008}\natexlab{}.
\newblock \showarticletitle{The {Role} of {Software} {Tools} in {Influencing} {Creative} {Problem} {Solving} in {Engineering} {Design} and {Education}}. \bibinfo{publisher}{American Society of Mechanical Engineers Digital Collection}, \bibinfo{pages}{999--1007}.
\newblock
\urldef\tempurl%
\url{https://doi.org/10.1115/DETC2006-99343}
\showDOI{\tempurl}


\bibitem[Sanders and Stappers(2008)]%
        {sanders_co-creation_2008}
\bibfield{author}{\bibinfo{person}{Elizabeth B.-N. Sanders} {and} \bibinfo{person}{Pieter~Jan Stappers}.} \bibinfo{year}{2008}\natexlab{}.
\newblock \showarticletitle{Co-creation and the new landscapes of design}.
\newblock \bibinfo{journal}{\emph{CoDesign}} \bibinfo{volume}{4}, \bibinfo{number}{1} (\bibinfo{date}{March} \bibinfo{year}{2008}), \bibinfo{pages}{5--18}.
\newblock
\showISSN{1571-0882}
\urldef\tempurl%
\url{https://doi.org/10.1080/15710880701875068}
\showDOI{\tempurl}
\newblock
\shownote{Publisher: Taylor \& Francis \_eprint: https://doi.org/10.1080/15710880701875068}.


\bibitem[Singh et~al\mbox{.}(2022)]%
        {singh_where_2022}
\bibfield{author}{\bibinfo{person}{Nikhil Singh}, \bibinfo{person}{Guillermo Bernal}, \bibinfo{person}{Daria Savchenko}, {and} \bibinfo{person}{Elena~L. Glassman}.} \bibinfo{year}{2022}\natexlab{}.
\newblock \showarticletitle{Where to {Hide} a {Stolen} {Elephant}: {Leaps} in {Creative} {Writing} with {Multimodal} {Machine} {Intelligence}}.
\newblock \bibinfo{journal}{\emph{ACM Transactions on Computer-Human Interaction}} (\bibinfo{date}{Feb.} \bibinfo{year}{2022}).
\newblock
\showISSN{1073-0516}
\urldef\tempurl%
\url{https://doi.org/10.1145/3511599}
\showDOI{\tempurl}
\newblock
\shownote{Just Accepted}.


\bibitem[Smith and Linsey(2011)]%
        {smith_three-pronged_2011}
\bibfield{author}{\bibinfo{person}{Steven~M. Smith} {and} \bibinfo{person}{Julie Linsey}.} \bibinfo{year}{2011}\natexlab{}.
\newblock \showarticletitle{A {Three}-{Pronged} {Approach} for {Overcoming} {Design} {Fixation}}.
\newblock \bibinfo{journal}{\emph{The Journal of Creative Behavior}} \bibinfo{volume}{45}, \bibinfo{number}{2} (\bibinfo{year}{2011}), \bibinfo{pages}{83--91}.
\newblock
\showISSN{2162-6057}
\urldef\tempurl%
\url{https://doi.org/10.1002/j.2162-6057.2011.tb01087.x}
\showDOI{\tempurl}
\newblock
\shownote{\_eprint: https://onlinelibrary.wiley.com/doi/pdf/10.1002/j.2162-6057.2011.tb01087.x}.


\bibitem[Summers-Stay et~al\mbox{.}(2023)]%
        {summers-stay_brainstorm_2023}
\bibfield{author}{\bibinfo{person}{Douglas Summers-Stay}, \bibinfo{person}{Clare~R. Voss}, {and} \bibinfo{person}{Stephanie~M. Lukin}.} \bibinfo{year}{2023}\natexlab{}.
\newblock \showarticletitle{Brainstorm, then {Select}: a {Generative} {Language} {Model} {Improves} {Its} {Creativity} {Score}}.
\newblock
\urldef\tempurl%
\url{https://openreview.net/forum?id=8HwKaJ1wvl}
\showURL{%
\tempurl}


\bibitem[Thrash et~al\mbox{.}(2014)]%
        {thrash_psychology_2014}
\bibfield{author}{\bibinfo{person}{Todd~M. Thrash}, \bibinfo{person}{Emil~G. Moldovan}, \bibinfo{person}{Victoria~C. Oleynick}, {and} \bibinfo{person}{Laura~A. Maruskin}.} \bibinfo{year}{2014}\natexlab{}.
\newblock \showarticletitle{The {Psychology} of {Inspiration}}.
\newblock \bibinfo{journal}{\emph{Social and Personality Psychology Compass}} \bibinfo{volume}{8}, \bibinfo{number}{9} (\bibinfo{year}{2014}), \bibinfo{pages}{495--510}.
\newblock
\showISSN{1751-9004}
\urldef\tempurl%
\url{https://doi.org/10.1111/spc3.12127}
\showDOI{\tempurl}
\newblock
\shownote{\_eprint: https://onlinelibrary.wiley.com/doi/pdf/10.1111/spc3.12127}.


\bibitem[Vaswani et~al\mbox{.}(2017)]%
        {vaswani_attention_2017}
\bibfield{author}{\bibinfo{person}{Ashish Vaswani}, \bibinfo{person}{Noam Shazeer}, \bibinfo{person}{Niki Parmar}, \bibinfo{person}{Jakob Uszkoreit}, \bibinfo{person}{Llion Jones}, \bibinfo{person}{Aidan~N Gomez}, \bibinfo{person}{Łukasz Kaiser}, {and} \bibinfo{person}{Illia Polosukhin}.} \bibinfo{year}{2017}\natexlab{}.
\newblock \showarticletitle{Attention is {All} you {Need}}. In \bibinfo{booktitle}{\emph{Advances in {Neural} {Information} {Processing} {Systems}}}, Vol.~\bibinfo{volume}{30}. \bibinfo{publisher}{Curran Associates, Inc.}
\newblock
\urldef\tempurl%
\url{https://proceedings.neurips.cc/paper/2017/hash/3f5ee243547dee91fbd053c1c4a845aa-Abstract.html}
\showURL{%
\tempurl}


\bibitem[Wooten and Ulrich(2017)]%
        {wooten_idea_2017}
\bibfield{author}{\bibinfo{person}{Joel~O. Wooten} {and} \bibinfo{person}{Karl~T. Ulrich}.} \bibinfo{year}{2017}\natexlab{}.
\newblock \showarticletitle{Idea {Generation} and the {Role} of {Feedback}: {Evidence} from {Field} {Experiments} with {Innovation} {Tournaments}}.
\newblock \bibinfo{journal}{\emph{Production and Operations Management}} \bibinfo{volume}{26}, \bibinfo{number}{1} (\bibinfo{year}{2017}), \bibinfo{pages}{80--99}.
\newblock
\showISSN{1937-5956}
\urldef\tempurl%
\url{https://doi.org/10.1111/poms.12613}
\showDOI{\tempurl}
\newblock
\shownote{\_eprint: https://onlinelibrary.wiley.com/doi/pdf/10.1111/poms.12613}.


\bibitem[Youmans and Arciszewski(2014)]%
        {youmans_design_2014}
\bibfield{author}{\bibinfo{person}{Robert~J. Youmans} {and} \bibinfo{person}{Thomaz Arciszewski}.} \bibinfo{year}{2014}\natexlab{}.
\newblock \showarticletitle{Design fixation: {Classifications} and modern methods of prevention}.
\newblock \bibinfo{journal}{\emph{AI EDAM}} \bibinfo{volume}{28}, \bibinfo{number}{2} (\bibinfo{date}{May} \bibinfo{year}{2014}), \bibinfo{pages}{129--137}.
\newblock
\showISSN{0890-0604, 1469-1760}
\urldef\tempurl%
\url{https://doi.org/10.1017/S0890060414000043}
\showDOI{\tempurl}
\newblock
\shownote{Publisher: Cambridge University Press}.


\bibitem[Yun et~al\mbox{.}(2021)]%
        {yun_collective_2021}
\bibfield{author}{\bibinfo{person}{Jinhyo~Joseph Yun}, \bibinfo{person}{Euiseob Jeong}, \bibinfo{person}{Sangwoo Kim}, \bibinfo{person}{Heungju Ahn}, \bibinfo{person}{Kyunghun Kim}, \bibinfo{person}{Sung~Deuk Hahm}, {and} \bibinfo{person}{Kyungbae Park}.} \bibinfo{year}{2021}\natexlab{}.
\newblock \showarticletitle{Collective {Intelligence}: {The} {Creative} {Way} from {Knowledge} to {Open} {Innovation}}.
\newblock \bibinfo{journal}{\emph{Science, Technology and Society}} \bibinfo{volume}{26}, \bibinfo{number}{2} (\bibinfo{date}{July} \bibinfo{year}{2021}), \bibinfo{pages}{201--222}.
\newblock
\showISSN{0971-7218}
\urldef\tempurl%
\url{https://doi.org/10.1177/09717218211005604}
\showDOI{\tempurl}
\newblock
\shownote{Publisher: SAGE Publications India}.


\bibitem[Zhu and Luo(2023)]%
        {zhu_generative_2023}
\bibfield{author}{\bibinfo{person}{Qihao Zhu} {and} \bibinfo{person}{Jianxi Luo}.} \bibinfo{year}{2023}\natexlab{}.
\newblock \showarticletitle{Generative {Transformers} for {Design} {Concept} {Generation}}.
\newblock \bibinfo{journal}{\emph{Journal of Computing and Information Science in Engineering}} \bibinfo{volume}{23}, \bibinfo{number}{4} (\bibinfo{date}{Jan.} \bibinfo{year}{2023}).
\newblock
\showISSN{1530-9827}
\urldef\tempurl%
\url{https://doi.org/10.1115/1.4056220}
\showDOI{\tempurl}


\end{thebibliography}

\newpage
\phantom{bumping things to a new page}
\newpage
\phantom{bumping things to a new page}
\newpage

\section{Appendices}
\appendix

\section{Using the Supermind Ideator API}

To run a move all one needs to provide is a query(). For example, to run our experimental Reflect move (a zero-shot prompt), the query is defined as:

\small
\begin{lstlisting}[language=Python, showstringspaces=false, breaklines=true]
def query(problem):
return """I have the following problem statement: {}. What am I missing? What else should I think about when formulating my need? Use the answers to these questions to create better problem statements.""".format(problem)
\end{lstlisting}
\normalsize

The prompt string uses a built-in Python method format to insert the given problem variable into the prompt before sending that off to GPT. This, we believe, makes creating, understanding, and adapting moves more accessible to individuals of all technical skill levels. For more advanced move design, we also support a range of optional parameters including:
\small
\begin{itemize}
    \item prompt(): a string that provides a standalone leading prompt that can be sent in addition to the query
    \item few\_shot(): a string that provides example input and output that should be given to GPT before submitting a query in order to quickly train desired behavior
    \item stop(): a string that provides a custom termination for GPT in order to inform the system when to stop
    \item finetune\_model(): a string that provides the name of a custom finetune model.(Note: This likely requires the use of a specific API key due to how OpenAI provides access to fine-tune models.)
    \item system\_message(): a string that can be provided at the beginning of a query to better tune chat-based model behavior (such as with ChatGPT or GPT-4)
\end{itemize}
\normalsize

\end{document}